\documentclass[10pt,twocolumn,letterpaper]{article}

\usepackage[pagenumbers]{cvpr} 

\definecolor{cvprblue}{rgb}{0.21,0.49,0.74}
\usepackage[pagebackref,breaklinks,colorlinks,allcolors=cvprblue]{hyperref}
\usepackage{lipsum}
\usepackage{algorithm}%
\usepackage{algorithmicx}%
\usepackage{algpseudocode}%
\usepackage{multirow}
\def\paperID{36} 
\def\confName{CVPR}
\def\confYear{2025}

\title{DEFT-VTON: Efficient Virtual Try-On with Consistent Generalised H-Transform}

\author{
Xingzi Xu\textsuperscript{1,2 *\textdagger}
\quad
Qi Li\textsuperscript{1}\thanks{Equal contribution.} \quad
Shuwen Qiu\textsuperscript{3}\thanks{Work done during internship at Amazon.}  \quad
Julien Han\textsuperscript{1} \quad
Karim Bouyarmane\textsuperscript{1} \\
\textsuperscript{1}Amazon \quad
\textsuperscript{2}Duke University \quad
\textsuperscript{3} University of California, Los Angeles (UCLA) \\
\textsuperscript{1}{\tt\small \{qlimz,hameng,bouykari\}@amazon.com} \\
\textsuperscript{2}{\tt\small xingzi.xu@duke.edu} \quad
\textsuperscript{3}{\tt\small janetqiu@cs.ucla.edu} \\
{\color{red}{\tt\small \url{https://deft-vton.github.io/}}}
}

\newtheorem{theorem}{{\bf Theorem}}

\newcommand{\mP}{\mathcal{P}}
\newcommand{\bX}{\mathbf{X}}
\newcommand{\bY}{\mathbf{Y}}
\newcommand{\bZ}{\mathbf{Z}}

\newcommand{\bI}{\mathbf{I}}
\newcommand{\bx}{\mathbf{x}}
\newcommand{\by}{\mathbf{y}}

\newcommand{\bfs}{\mathbf{s}}
\newcommand{\bfh}{\mathbf{h}}
\newcommand{\bH}{\mathbf{H}}
\newcommand{\bW}{\mathbf{W}}
\newcommand{\bR}{\mathbb{R}}

\begin{document}
 \maketitle
 \begin{abstract}
Diffusion models enables high-quality virtual try-on (VTO) with their established image synthesis abilities. Despite the extensive end-to-end training of large pre-trained models involved in current VTO methods, real-world applications often prioritize limited training and inferencing/serving/deployment budgets for VTO. To solve this obstacle, we apply Doob’s h-transform efficient fine-tuning (DEFT) for adapting large pre-trained unconditional models for downstream image-conditioned VTO abilities. DEFT freezes the pre-trained model’s parameters and trains a small h-transform network to learn a conditional h-transform. The h-transform network allows to train only ~$1.42$\% of the frozen parameters, compared to baseline ~$5.52$\% in traditional parameter-efficient fine-tuning (PEFT).
To further improve DEFT’s performance, and decrease existing models’ inference time, we additionally propose an adaptive consistency loss. Consistency training distills slow but performing diffusion models into a fast one while retaining performances by enforcing consistencies along the inference path. Inspired by constrained optimization, instead of distillation, we combine the consistency loss and the denoising score matching loss in a data-adaptive manner for fine-tuning existing VTO models at a low cost. Empirical results show proposed DEFT-VTON method achieves SOTA performances on VTO tasks, as well as a number of function evaluations (denoising steps) as low as 15, while maintaining competitive performances.
\end{abstract}    
 \section{Introduction}
\label{sec:intro}

\begin{figure*}[t]
    \centering
    \includegraphics[width=\textwidth, trim={0in 3.9in 0in 0in},clip]{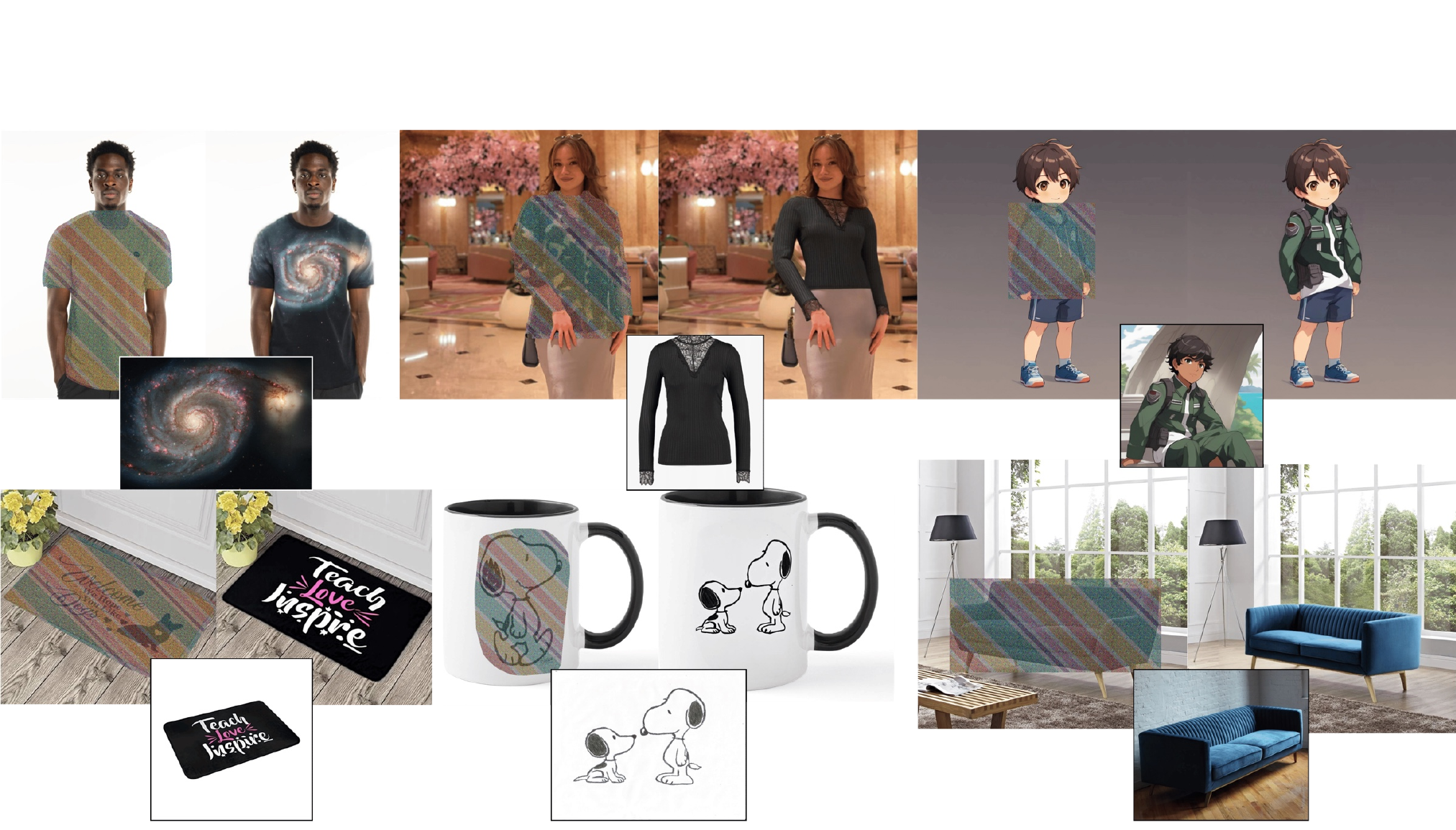}
    \caption{DEFT-VTON achieves realistic results across diverse application domains.}
    \label{fig:qualititive}
\end{figure*}
\vspace{-0in}

Virtual try-on (VTO) technology has emerged as a transformative solution in the e-commerce fashion and general image editing industries, addressing the critical gap between online shopping convenience and the traditional in-store fitting experience \citep{seyfioglu2024diffusechooseenrichingimage, song2024imagebasedvirtualtryonsurvey,seyfioglu2023dreampaint}. VTO allows customers to visualize how clothing items would look on themselves without physical interaction with the garments, revolutionizing the online shopping experience.
Virtual try-on technology has evolved significantly with the emergence of denoising diffusion models. Denoising diffusion models are a class of expressive generative models that progressively convert noise to realistic data \citep{ho2020denoisingdiffusionprobabilisticmodels,song2021scorebasedgenerativemodelingstochastic, rombach2022high}. 

Large-scale denoising diffusion models have achieved unparalleled success on unconditional image synthesis \citep{dhariwal2021diffusionmodelsbeatgans}. Unconditional diffusion models approximate the score of the underlying distribution $s^{\theta}(t,\mathbf{x})\approx \nabla_{\mathbf{x}}\log p_t(\mathbf{x})$. For conditional generations regarding a constraint $\mathbf{y}$, we can estimate the conditional score based on the unconditional one through the Bayes’ theorem:

\begin{equation}
    \nabla_{\mathbf{x}}\log p_t(\mathbf{x}|\mathbf{Y}=\mathbf{y})\approx s^{\theta}(t,\mathbf{x})+\nabla_{\mathbf{x}}\log p_t(\mathbf{Y}=\mathbf{y}|\mathbf{x}),
\end{equation}

where $\nabla_{\mathbf{x}}\log p_t(\mathbf{Y}=\mathbf{y}|\mathbf{x})$ represents the guidance guiding the denoising process towards the conditioned results. Existing methods explore training-free sampling of conditioned processes with an unconditioned model as well as fine-tuning methods. Although training-free methods do not suffer from extensive retraining and fine-tuning of the unconditional model, they suffer from a slow sampling process rooting in the expensive estimation of the guidance and slower convergence rates than their unconditional counterpart \citep{chung2024diffusionposteriorsamplinggeneral,song2023pseudoinverseguided}. On the other hand, fine-tuning methods on large unconditional methods require impractical training costs, as well as large amounts of paired data points \citep{mardani2023variationalperspectivesolvinginverse, ho2022classifierfreediffusionguidance, batzolis2021conditionalimagegenerationscorebased, liu2023i2sbimagetoimageschrodingerbridge}. 
Doob’s h-transform efficient fine-tuning (DEFT) employs a large-scale pre-trained unconditional diffusion model, which can achieve high-quality unconditional generation, but is computationally intractable to fine-tune \citep{denker2024deftefficientfinetuningdiffusion}. DEFT instead suggests learning a small-scale network $h$ to directly approximate the guidance $\nabla_{\mathbf{x}}\log p_t(\mathbf{Y}=\mathbf{y}|\mathbf{x})$. DEFT freezes the weights of the unconditional model, and achieves state-of-the-art (SOTA) perceptual qualities on tasks such as image reconstruction while achieving speedups \citep{denker2024deftefficientfinetuningdiffusion, kawar2022denoisingdiffusionrestorationmodels}.

Despite their superior generation capabilities, diffusion models are slower than VAEs and GANs because of their large number of function evaluations (NFEs), even with highly optimized samplers \citep{song2022denoisingdiffusionimplicitmodels, zhang2023fastsamplingdiffusionmodels,karras2022elucidatingdesignspacediffusionbased}. Meanwhile, distilling, pre-training, and regularizing with consistency losses along inference paths have provided improved generation qualities and speeds \citep{song2023consistencymodels,song2023improvedtechniquestrainingconsistency, daras2023consistentdiffusionmodelsmitigating, kim2024consistencytrajectorymodelslearning, geng2024consistencymodelseasy}. 

We propose DEFT-VTON to train an adaptor that directs unconditional generations of a pre-trained diffusion model to that of a VTO conditioned one. In particular, we propose an adaptive loss that balances the consistency loss and the DEFT loss on the fly, achieving SOTA performances on VTO tasks, while using only a limited computational budget. Empirical results suggest adding the consistency loss speeds up the VTO inference up to 40\% while maintaining the same performances.

\begin{figure*}[t]
    \centering
    \includegraphics[width=0.5\textwidth]{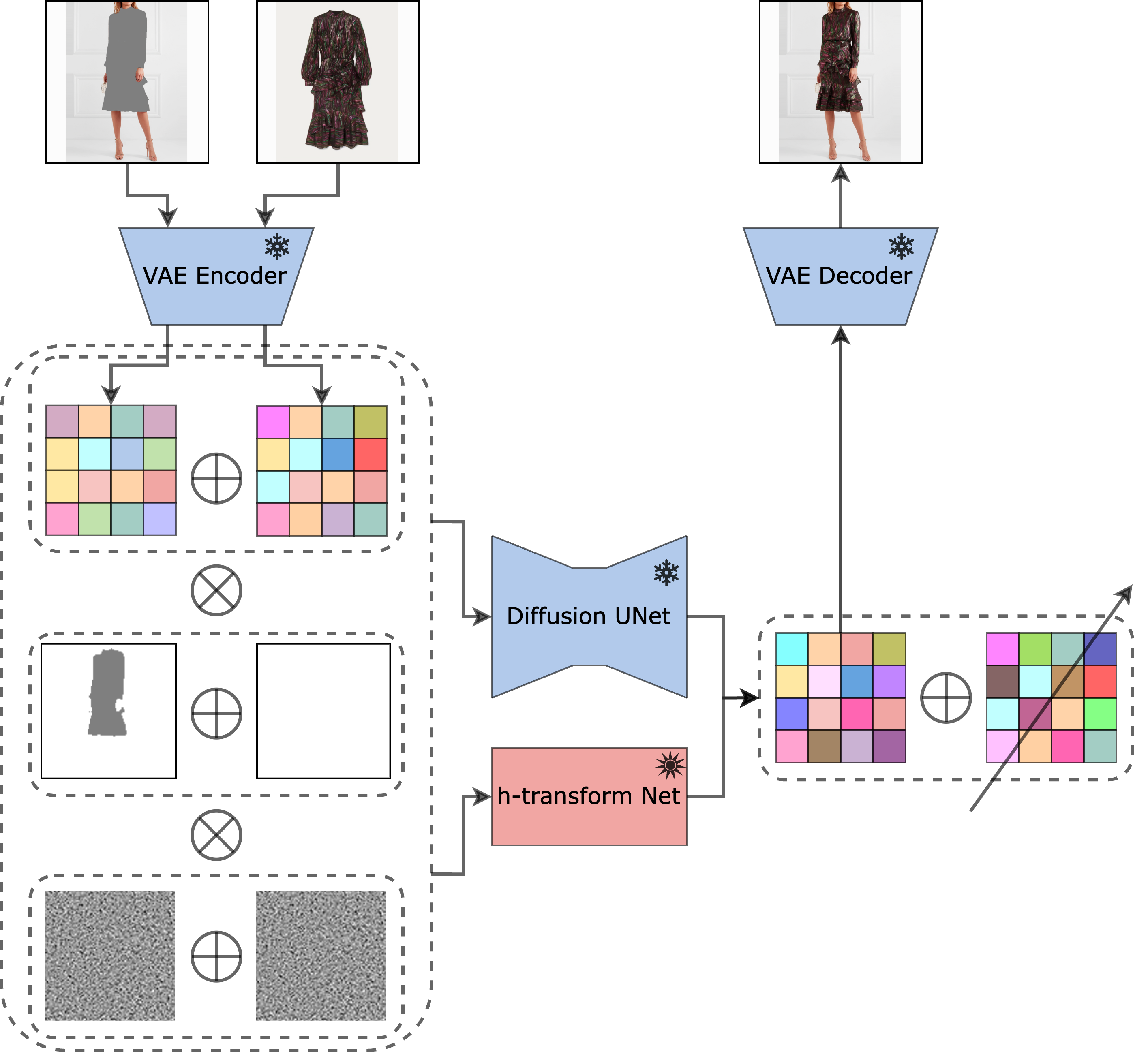}
    \caption{We train an efficient h-transform network to perform virtual try-on tasks. $\oplus$ indicates combination along spatial dimension, $\otimes$ indicates combination along channel dimension, $\nearrow$ indicates discarding the specified part of the tensor, blue indicates frozen parameters, red indicates trainable parameters.}
    \label{fig:main}
\end{figure*}

 \section{Related works}
\label{sec:related}

\paragraph{Virtual try-on}
Given an image of a person and a target garment, virtual try-on (VTO) methods aim to generate an image of the person wearing the target garment, while preserving the garment's fine-grained details and blending naturally into the surrounding context \cite{seyfioglu2024diffusechooseenrichingimage}. Traditional methods combine pose, body shape, and other garment-agnostic person representations with images of the target garment for generation \citep{han2018vitonimagebasedvirtualtryon,bai2022singlestagevirtualtryon,choi2021viton,Gou_2023,karras2020analyzingimprovingimagequality,lee2022highresolutionvirtualtryonmisalignment,lewis2021tryonganbodyawaretryonlayered,xie2023gpvtongeneralpurposevirtual,xue2022dccfdeepcomprehensiblecolor,10205434, chong2024catvtonconcatenationneedvirtual, kim2023stablevitonlearningsemanticcorrespondence, chong2024catvtonconcatenationneedvirtual, qi2024ditvton, qi2024posevton, julien2024instructvton}. 

\paragraph{Consistency models}
Consistency models accelerate and even avoid the iterative sampling process of diffusion models, supporting fast one-step generation by design, while still allowing multi-step sampling to trade compute for sample quality \citep{song2023consistencymodels}. Given a probability flow ODE that smoothly converts data to noise, consistency models learn to map any point on the ODE trajectory to the data for generative modeling. As all points along the ODE trajectory are trained to map to the same data point, these mappings are called consistency models. \citep{kim2024consistencytrajectorymodelslearning,li2024bidirectionalconsistencymodels,song2023improvedtechniquestrainingconsistency,yang2024consistencyflowmatchingdefining} expands on the idea, achieving better generation qualities while maintaining low NFEs. Remaining consistent along generation paths have also proved beneficial to avoiding sampling errors, and improves generation performances~\citep{daras2023consistentdiffusionmodelsmitigating}.

\paragraph{Conditional diffusion models}
Conditional generation with diffusion models can be generally divided into training-free and training methods. Training-free methods evaluate score guidance at inference time, guiding unconditional generations towards that of a conditional one \citep{chung2024diffusionposteriorsamplinggeneral,kawar2022denoisingdiffusionrestorationmodels,mardani2023variationalperspectivesolvinginverse,song2023pseudoinverseguided,chung2024improvingdiffusionmodelsinverse}. Despite these methods do not require any training computations, they require expensive inference time evaluations, hindering their use in large scale and high resolution VTO tasks. \citep{ho2022classifierfreediffusionguidance,mardani2023variationalperspectivesolvinginverse} fine-tunes the pre-trained model for better conditional performances. However, such fine-tuning can be unpractical for large scale pre-trained models. \citep{dhariwal2021diffusionmodelsbeatgans} trains a small-scale classifier and use a guidance based on the gradient with respect to the classifier at inference time. However, this classifier must be trained on noisy data so it is generally not possible to plug
in a pre-trained classifier, and the training and sampling resembles that of a GAN, leading to an inflated performance on the evaluation criteria~\citep{ho2022classifierfreediffusionguidance}. DEFT trains an adaptor network extra to the pre-trained model to approximate a Doob's h-transform function, efficiently achieving SOTA perceptual quality and inference speeds on image reconstruction tasks.
 \section{Method}
\label{sec:method}
We now introduce our proposed DEFT-VTON method. We first give brief reviews to a baseline VTO model, DEFT, and consistency model, and then combine with constrained optimization to propose an adaptive loss used for training DEFT-VTON.

\subsection{Baseline model}
\begin{figure}[hbt!]
    \centering
    \includegraphics[width=0.5\textwidth]{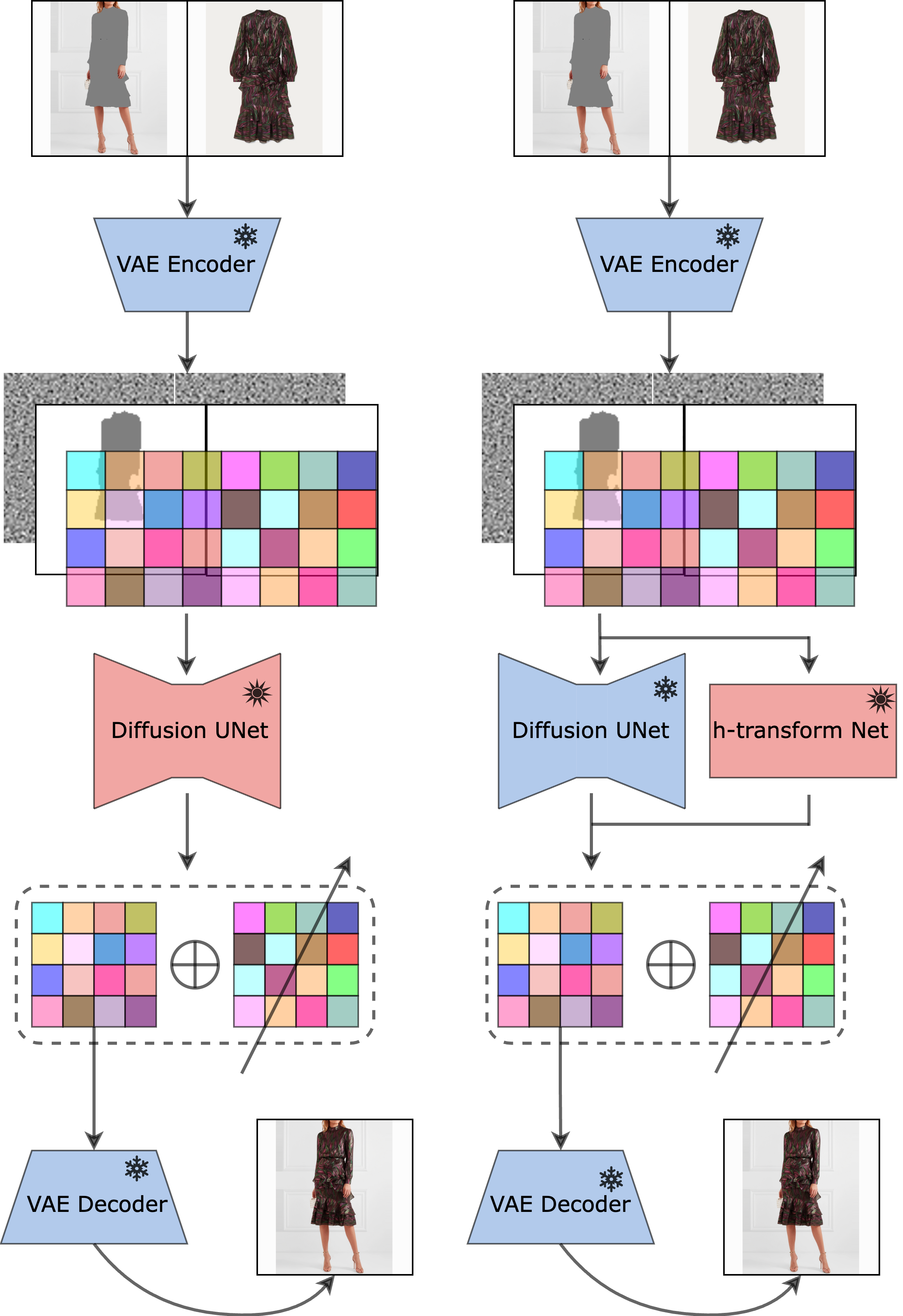}
    \caption{Structure of DEFT-VTON (right) compared to baseline PEFT architecture (left). While the baseline PEFT trains ~5.51\% percent of the backbone network, we freeze the backbone completely and train an auxiliary network with ~1.42\% percent of the backbone network parameters.}
    \label{fig:catvton}
\end{figure}

We use a Latent Diffusion baseline VTO architecture as base model, for similar architectures see~\citep{rombach2022high,seyfioglu2024diffusechooseenrichingimage,chong2024catvtonconcatenationneedvirtual}. The training process takes a target image $\bI_p\in\bR^{3\times H\times W}$, a binary garment reference image $\bI_g\in\bR^{3\times H\times W}$, and a mask image $\bI_M\in\bR^{H\times W}$. We define the garment agnostic image as:
\begin{equation}
    \bI_a=\bI_p\circ \bI_M,
\end{equation}
where $\circ$ represents the element-wise product. Then we apply pre-trained VAE encoder $\varepsilon$ to the garment agnostic image $\bI_a$ and the garment reference image (in the form of in-shop garment or used in-scene) $\bI_g$:
\begin{align}
    \bZ_a &= \varepsilon(\bI_a),\\
    \bZ_g &= \varepsilon(\bI_g),
\end{align}
where $\bZ_a,\bZ_g \in \bR^{C_{\bZ}, H_{\bZ}, W_{\bZ}}$, with $H_{\bZ}, W_{\bZ} \ll H, W$. We interpolate the mask $\bI_M$ to match the latent space size and get $\bI_m\in\bR^{C_{\bZ}\times H_{\bZ}\times W_{\bZ}}$. Next, we join $\bZ_a,\bZ_g$ along the spatial dimension to get $\bZ_c\in\bR^{C_{\bZ}\times H_{\bZ}\times 2W_{\bZ}}$, and join the mask $I_m$ with an all-zero matrix of the same size to create $\bI_{mc}\in\bR^{C_{\bZ}\times H_{\bZ}\times 2W_{\bZ}}$:
\begin{equation}
\label{eqn:ZI}
    \bZ_c = \bZ_a \oplus\bZ_g,\,\bI_{mc} = \bI_m \oplus \mathbb{O},
\end{equation}
where $\oplus$ represents joining along the spatial dimension, and $\mathbb{O}$ represents the all-zero matrix. At the beginning of the denoising process, we sample a Gaussian noise matrix $\bW\sim\mathcal{N}(0,\mathbb{I})$ of the same shape as $\bZ_c$ and $\bI_{mc}$ to join with them along the channel dimension and get
\begin{align}
    \bX_T = \bZ_c\otimes\bI_{mc}\otimes\bW,
\end{align}
where $\otimes$ represents joining along the channel dimension. $\bX_T$ goes through the UNet $\bfs$ to predict $\bX_{T-1}$, and iterate for a large number $T$ times to predict a clean $\bX_0$. In particular, we have
\begin{align}
    \bX_{t-1} = \bfs(\bZ_c\otimes\bI_{mc}\otimes\bX_t, t).
\end{align}
Finally, we split $\bX_0\in\bR^{C_{\bZ}\times H_{\bZ}\times 2W_{\bZ}}$ along the spatial dimension to get the latent VTO result $\bX_0^{\text{VTO}}\in\bR^{C_{\bZ}\times H_{\bZ}\times W_{\bZ}}$, and use the pre-trained VAE decoder $\mathbb{D}$ to transform back to the image space, getting our VTO image prediction $\hat{I}^{\text{VTO}}\in\bR^{3\times H\times W}$.

The baseline model freezes the weights of the VAE, and performs a parameter-efficient training of the diffusion U-Net. In particular, the model training consists of first adding various levels of noises to the encoded target image $\varepsilon(\bI_p)$ and then predicting the noises added \citep{salimans2022progressivedistillationfastsampling}.

\subsection{DEFT}
We now give a brief review to conditioning diffusions with the h-transform. Starting with a forward stochastic differential equation (SDE) transforming a data distribution $\mP_0=p_{\text{data}}$ to a Gaussian distribution $\mP_T=\mathcal{N}(0,\mathbb{I})$:
\begin{equation}
    d\bX_t=f_t(\bX_t)dt+\sigma_td\bW_t,\bX_0\sim\mP_0,
\end{equation}
where the drift $f_t$ and the diffusion $\sigma_t$ are given explicitly. Specifically, in DDPM discretizations, we have $f_t(\bX_t)=-\frac{\beta(t)}{2}\bX_t$, and $\sigma_t=\sqrt{\beta(t)}$, with $\beta(t)$ explicitly given. Under mild assumptions, there exists a reverse SDE with drift $b_t$ that transforms $\mP_T$ back to $\mP_0$:
\begin{align}
\label{eqn:reverse}
    d\bX_t &= b_t(\bX_t)dt+\sigma_td\bW_t,\bX_T\sim\mP_T,\\
    b_t(\bX_t) &= f_t(\bX_t)-\sigma_t^2\nabla_{\bX_t}\log p_t(\bX_t).
\end{align}
Unconditional diffusion models trains by approximating the score function $\bfs(\bX_t,t)=\nabla_{\bX_t}\log p_t(\bX_t)$, which can be cheaply evaluated for certain drift and diffusion functions \cite{song2022denoisingdiffusionimplicitmodels, karras2022elucidatingdesignspacediffusionbased}. 

\begin{figure}[hbt!]
    \centering
    \includegraphics[width=0.5\textwidth]{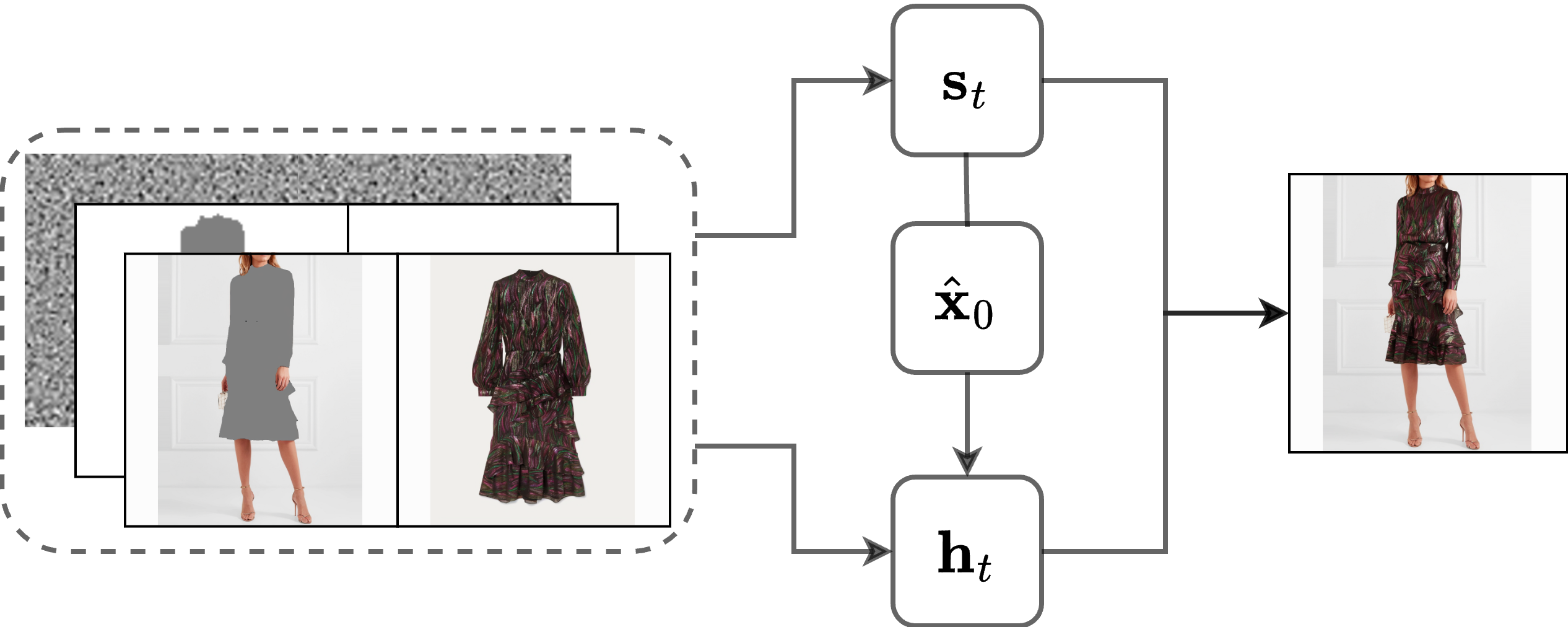}
    \caption{H-transform network takes the same inputs as the diffusion U-Net, and guides the unconditional generator towards VTON results.}
    \label{fig:deft}
\end{figure}

Conditional diffusion models want to condition the reverse SDE on a particular observation. Doob's h-transform provides a mathematical tool for guiding an SDE to hit a event at a given time \citep{heng2022simulatingdiffusionbridgesscore,Rogers_Williams_2000}:
\begin{theorem}[Doob's h-transform \citep{denker2024deftefficientfinetuningdiffusion, Rogers_Williams_2000}]
    Consider the equation \ref{eqn:reverse}, the conditioned process $\bX_t|\bX_0\in B$ is a solution to:
    \[
    \begin{cases}
        d\bH_t=(b_t(\bH_t)-\sigma_t^2\nabla_{\bH_t}\log p_{0|t}(\bX_0\in B|\bH_t))dt+\sigma_td\bW_t, \\
        \bH_t\sim\mP_T,
    \end{cases}
    \]
    where we represent unconditional processes with $\bX_t$ and conditional processes with $\bH_t$, $b_t(\bH_t) = f_t(\bH_t)-\sigma_t^2\nabla_{\bH_t}\log p_t(\bH_t)$ and $\mathbb{P}(\bX_0\in B)=1$.
\end{theorem}

In the case of VTO, we are interested in inpainting tasks, and conditioning on $\bY=\bZ_c\otimes \bI_{mc}$, where $\bZ_{c}$ and $\bI_{mc}$ are the latent reference and the interpolated mask, as defined in equation \ref{eqn:ZI}. We aim to sample from the posterior $p(\bX=x_0|\bY=\by)$. The h-transform admits a denoising score matching objective given as follows:
\begin{theorem}[DSM for generalised h-transform \citep{denker2024deftefficientfinetuningdiffusion}]
\label{thm:dsm}
    For a given condition $\by$, let $\mathbb{Q}$ be the path measure of the conditional SDE
    \begin{align}
        d\bH_t = (f_t(\bH_t)-\sigma_t^2(\nabla_{\bH_t}\log p_t(\bH_t)))dt+\sigma_t d\bW_t,
    \end{align}
    where $\bH_T\sim Q_T^{f_t}[p(\bx_0|\by)]=\int p_{T|0}(\bx|\bx_0)p(\bx_0|\by)d\bx_0$ converges to the Gaussian distribution $\mathcal{N}(0,I)$ exponentially w.r.t. $T$ for VP-SDE. The h-transform then admits a denoising score matching (DSM) objective:
    \begin{align}
        &\bfh^*=\arg\min_{\bfh\in\mathcal{H}}\mathcal{L}_{DSM}^{\by}(\bfh),\\
        &\mathcal{L}_{DSM}^{\by}(\bfh) \equiv \mathbb{E}_{\bX_0\sim p(\bx_0|\by),t\sim \text{Unif}(0,T),\bH_t\sim p_{t|0}(\bx_t|\bx_0)}&\\
        &[\|(h_t(\bH_t)+\nabla_{\bH_t}\log p_t(\bH_t))-\nabla_{\bH_t}\log p_{t|0}(\bH_t|\bX_0)\|^2]
    \end{align}
\end{theorem}
Theorem \ref{thm:dsm} induces that the generalized h-transform $\bfh_t(\bx,\by)=\nabla_{\bx}\log p_t(\by|\bx)$ can be approximated by solving the minimization problem 
\begin{equation}
\label{eqn:minh}
    \min_{\bfh\in\mathcal{H}}\mathbb{E}_{\by\sim\bY}[\mathcal{L}_{DSM}^{\by}(\bfh)].
\end{equation}
With a DDPM discretization of the SDE and a epsilon-matching formulated pre-trained model $\bfs_t^{\theta^*}$, this minimization problem \ref{eqn:minh} reduces to 
\[
\begin{cases}
\label{loss:h}
    \min_{h\in\mathcal{H}}\mathbb{E}_{(\bx_0,\by),\epsilon,t}[\|(h(\bH_t,\by)+\bfs_t^{\theta^*}(\bH_t,t))-\epsilon\|^2],\\
    \bH_t = \sqrt{\Bar{\alpha}_t}\bX_0+\sqrt{1-\Bar{\alpha}_t}\epsilon,\\
    (\bx_0,\by)\sim p(\bX_0,\bY),\\
    \Bar{\alpha}_t=\exp(-\int_0^T\beta(s)ds),\\
    \epsilon\sim\mathcal{N}(0,\bf{I}).
\end{cases}
\]
We use this minimization formulation for learning the VTO h-transform network to guide an unconditional diffusion model.

\subsection{Consistency model}
Consistency training accelerates diffusion models while maintaining similar performances \citep{geng2024consistencymodelseasy}. At the same time, enforcing consistencies along inference trajectories reduces shift between the training and the sampling distribution of diffusion models and improves the performances \citep{daras2023consistentdiffusionmodelsmitigating}. Different from the iterative generation of diffusion models, consistency models are defined as $\bfs:(\bx_t,t)\rightarrow\bx_{\eta}$, where $\eta$ is a fixed small number, and are trained to predict the data point in one step \citep{song2023consistencymodels}. Specifically, consistency models satisfy the self-consistency property:
\begin{equation}
    \bfs(\bx_t,t)=\bfs(\bx_{t'},t'),\forall t,t'\in[\eta,T].
\end{equation}

\begin{figure}[hbt!]
    \centering
    \includegraphics[width=0.5\textwidth]{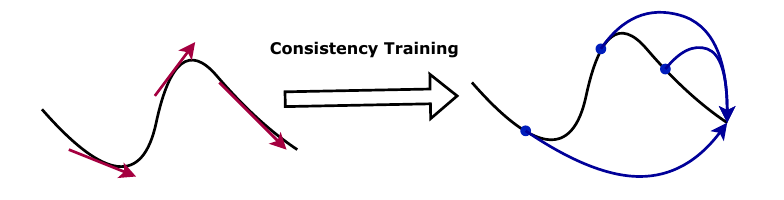}
    \caption{Consistency finetuning reduces number of function evaluations of diffusion models in the sampling process.}
    \label{fig:cm}
\end{figure}

Existing works distill a pre-trained diffusion model into a consistent model, greatly accelerating the generation process \citep{luo2023lcmlorauniversalstablediffusionacceleration, luo2023latentconsistencymodelssynthesizing}. Initializing the distilled parameters with $\theta$, we maintain a target model with parameters $\theta^-$, which is updated with the exponential moving average (EMA) of $\theta$, defined as $\theta^-=\mu\theta^-+(1-\mu)\theta$. The distillation model is trained by minimizing the consistency loss:
\begin{equation}
    \mathcal{L}(\theta,\theta^-)=\mathbb{E}_{\bx,n}[d(\bfs_{\theta}(\bx_{t_{n+1}},t_{n+1}),\bfs_{\theta^-}(\bx_{t_{n}},t_{n}))],
\end{equation}
where $d(\cdot,\cdot)$ is a chosen metric. We follow \citep{song2023improvedtechniquestrainingconsistency} and use the pseudo-Huber loss. In our case, as we do not perform distillations, and have an h-transform network with drastically different structures to the pre-trained diffusion model, we instead define the following training streamline:
\[
\begin{cases}
\label{loss:cm}
        \min_{\bfh\in\mathcal{H}}\mathbb{E}_{\bx,t}[d(\bfs_{\theta}(\bx_t,t)+\bfh(\bx_t,t),\bfs_{\theta}(\bx_{t'},t')+\bfh(\bx_{t'},t'))],\\
        \bx_t = \sqrt{\Bar{\alpha}_t}\bx_0+\sqrt{1-\Bar{\alpha}_t}\epsilon,\\
        \epsilon\sim\mathcal{N}(0,\bf{I}).
\end{cases}
\]
We follow the noising schedule and weighting function similar to~\citep{song2023improvedtechniquestrainingconsistency}. Sampling from a consistency model is similar to sampling from a DDIM, where we first predict the clean data at each step, and add slightly less noises. We refer to algorithm 1 of \cite{song2023consistencymodels} for similar sampling process.

\subsection{Adaptive loss}
We now combine the DEFT training in \ref{loss:h} and consistency training in \ref{loss:cm} adaptively, drawing ideas from constrained optimization. We first perform DEFT training in isolation for a performant h-transform network, using architectures as shown in Figure \ref{fig:main}. Afterwards, we finetune the h-transform network with the consistency loss for better sampling while retaining the VTO abilities. This problem can then be formulated as:

\begin{align}
    &\min_{h\in\mathcal{H}} \mathbb{E}_{\by\sim\bY}[\mathcal{L}_{CM}^{\by}(\bfh)+\mathcal{L}_{DSM}^{\by}(\bfh)],\\
    &\text{such that: }\\
    &\mathcal{L}_{CM}^{\by}(\bfh)=\\
    &\mathbb{E}_{(\bX_0),\epsilon,t}[\|\bfh(\bH_t,\by)+\bfs_t^{\theta^*}(\bH_t)-\epsilon\|^2]\leq\\
    &\mathbb{E}_{(\bX_0),\epsilon,t}[\|\bfh^{DEFT}(\bH_t,\by)+\bfs_t^{\theta^*}(\bH_t)-\epsilon\|^2],
\end{align}

where $\bfh^{DEFT}$ is the minimizer of the DEFT objective $\mathbb{E}_{\by\sim\bY}\mathcal{L}_{DSM}^{\by}(\bfh)$. Vanilla method to solve this problem at scale is minimizing $\mathcal{L}_{CM}+\lambda\mathcal{L}_{DSM}$. However, this vanilla method risks losing VTO abilities, as the network is able to find easy local minima with small consistency loss $\mathcal{L}_{CM}$. We generalize on this loss and propose to minimize
\begin{align}
\mathcal{L}_{adaptive}&=\lambda_{DSM1}\max(\mathcal{L}_{DSM},b_1)\\
&+\lambda_{DSM2}\min(\mathcal{L}_{DSM},b_2)+\lambda_{CM}\mathcal{L}_{CM},
\end{align}
where $\lambda_{CM}$, $\lambda_{DSM1}$, $\lambda_{DSM2}$, $b_1$, and $b_2$ are constants. The h-transform network trains for better VTO abilities when the VTO performance on a data point is lackluster, while focusing on the consistency objective when the VTO performance is good. Empirical results show that this adaptive loss helps preserve VTO performances while accelerating the sampling process.

\begin{algorithm}
\caption{Adaptive training algorithm for $\mathbf{x}_0$ prediction}\label{alg:adaptiveTraining}
\begin{algorithmic}[1]
    \State \textbf{Input:} pre-trained encoder $\varepsilon$, pre-trained score network $\bfs$, initialized h-transform network $\bfh$, garment agnoistic images, mask images, reference images, and target images $\{\bI_a^{(i)}, \bI_M^{(i)}, \bI_g^{(i)}, \bI_p^{(i)}\}_{i=1}^N$, coefficients $\lambda_{DSM1},\lambda_{DSM2}$, $\lambda_{CM}$, threshold $b_1$, $b_2$.
    \State Freeze weights of $\varepsilon$ and $\bfs$
    \While{Not converged}
        \For{$i \in 1,\dots, N$}
        \State Interpolate image space mask $\bI_M$ to get latent space mask $\bI_m$.
        \State Encode input $\bx_0^{(i)}=(\varepsilon(\bI_a)\oplus\varepsilon(\bI_g))\otimes(\bI_m\oplus\mathbb{O})\otimes(\varepsilon(\bI_p))$
        \State Sample noise level $\{t_j\}_{j=1}^{N_t}\sim\text{Unif}(0,1)$.
        \State Sample noise $\mathbf{W}\sim\mathcal{N}(0,\bI)$.
        \State Inject noise $\{\bx_{t_j}^{i}=\sqrt{\Bar{\alpha}_{t_j}}\bx_0^{(i)}+\sqrt{1-\Bar{\alpha}_{t_j}}\mathbf{W}\}_{j=1}^{N_t}$.
        \State Infer $\{\Hat{\bx}_0^{(i,j)}=\bfs(t,\bx_{t_j}^{(i)})+h(t,\bx_{t_j}^{(i)})\}_{j=1}^{N_t}$.
        \State Compute $\mathcal{L}_{DSM}=\frac{1}{N_t}\sum_{j=1}^{N_t}\text{P-Huber}(\Hat{\bx}_0^{(i,j)}-\bx_0^{(i)})$, $\mathcal{L}_{CM}=\frac{1}{N_t-1}\sum_{j=1}^{N_t-1}\text{P-Huber}(\Hat{\bx}_0^{(i,j)}-\Hat{\bx}_0^{(i,j+1)})$.
        \State Return loss $\lambda_{DSM1}\max(\mathcal{L}_{DSM},b_1)+\lambda_{DSM2}\min(\mathcal{L}_{DSM},b_2)+\lambda_{CM}\mathcal{L}_{CM}$ for optimization.
        \EndFor
    \EndWhile
    \State \textbf{Return:} trained h-transform network $\bfh$.
\end{algorithmic}
\end{algorithm}

 \section{Experiment}
\label{sec:experiment}

\begin{figure*}[t]
    \centering
    \includegraphics[width=1.0\textwidth]{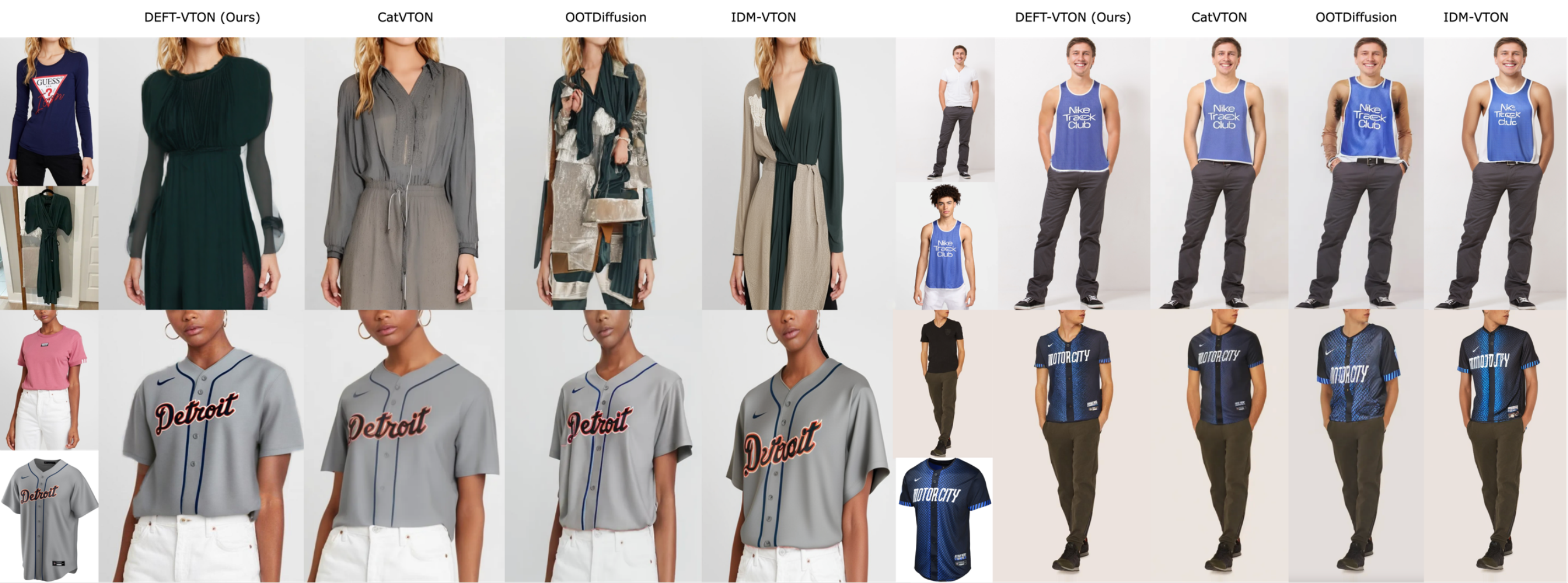}
    \caption{Comparison of our DEFT-VTON (15 sampling steps) with CatVTON (50 sampling steps) \citep{chong2024catvtonconcatenationneedvirtual}, OOTDiffusion (20 sampling steps) \citep{xu2024ootdiffusion}, and IDM-VTON (30 sampling steps) \citep{choi2024improving}.}
    \label{fig:adaptiveResult}
\end{figure*}

\subsection{Datasets}
\textbf{Virtual try-on test dataset} For testing, we conduct experiments using the test-split of the widely recognized public dataset VITON-HD~\citep{choi2021viton}
to compare against existing SOTA methods on blending image conditions into reference images in the VTO task. The VITON-HD dataset test-split is composed of pairs of reference garment images,  source images, and preprocessed mask and pose images. The source image serves as the ground-truth label, which is composed of the model wearing the item from the reference garment image. 
The VITON-HD test split includes 2,032 pairs of upper-body garment in front-facing poses. For our testing experiments, we reuse the preprocessed mask images provided in the dataset. 

\textbf{Virtual try-all dataset} In order to perform the try-on task beyond the front-facing garment try-on and commonly used garment categories like shirts and pants, as is in VITON-HD, we use a Virtual Try-All (VTA) dataset with diverse clothing and product categories, as well as novel object scenarios. The dataset includes 1) Expanded clothing categories, including jackets, pajamas, and more; 2) Clothing and non-clothing. Clothing include bags, hats, and shoes, non-clothing include jewelry, toys, and more. 3) More complicated source and reference images. 
Unlike existing datasets, reference images in the VTA dataset exhibits variety of object scenarios; for instance, the garment can be shown flat (lay-flat) instead of on a model in 3D form (also known as ghost-mannequin), it can also be shown being worn (on-figure) or not (off-figure).

\textbf{Data cleaning and preparation} To pick the right image pairs, we use a classifier to check if the source image is displaying the product in the reference image. 
To generate accurate masks, we first apply an object detection model to generate the bounding box area of the product in the source image, and then use the bounding box to prompt a segmentation model
to create the inpainting mask for the source image.



\subsection{Implementation details}
We use a Latent Diffusion model as our pre-trained model backbone.
The model consists of a pre-trained autoencoder and score diffusion U-Net. We freeze these networks and train an h-transform network .
During inference, we use a DDIM sampling method adapted to consistency models (for similar methods, see e.g.~\citep{song2023consistencymodels}). We perform training on 8 NVIDIA H100 GPUs.

\subsection{Evaluation metrics}
The evaluation is conducted in a paired and unpaired setting. In the paired setting, the original garment in the source image is the same as the garment provided in the reference image. The source image is served as the ground truth image. We use SSIM~\citep{wang2004image} and LPIPS~\citep{zhang2018unreasonable} to compare the ground truth with the generated results. In an unpaired image, the garment in the source image is a different garment from the referent image. We compute the distribution distance between the source image and the generated image with FID~\citep{zhang2018unreasonable} and KID~\citep{binkowski2018demystifying} score. We follow the implementation in~\citep{morelli2023ladi}. 

\subsection{Qualitative results}
\begin{figure*}[t]
    \centering
    \includegraphics[width=1.0\textwidth]{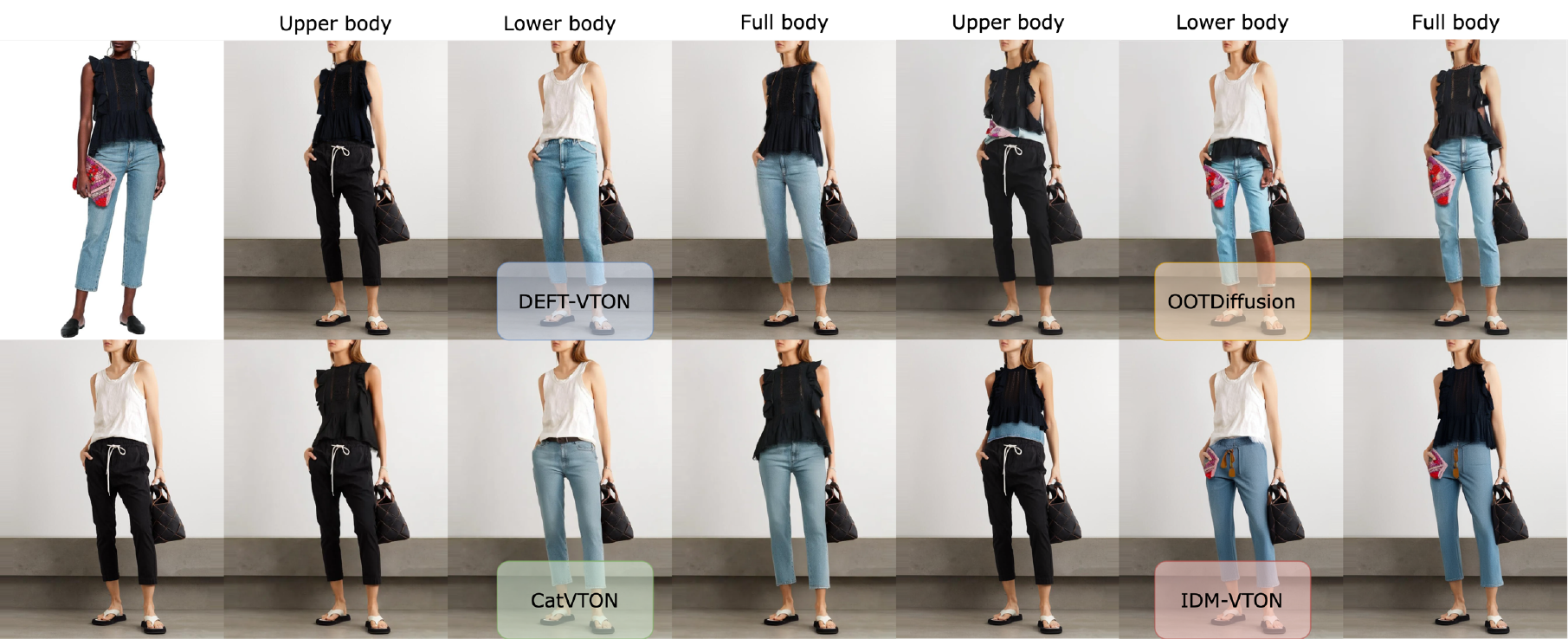}
    \caption{Comparison of DEFT-VTON with SOTA models on complex multi-garment try-on task.}
    \label{fig:SOTA}
\end{figure*}

We provide four semantic examples in Figure \ref{fig:adaptiveResult} on DEFT-VTON’s performances with only 15 sampling steps, and compare to that of SOTA models, CatVTON with 50 sampling steps, OOTDiffusion with 20 sampling steps, and IDM-VTON with 30 sampling steps \citep{xu2024ootdiffusion, chong2024catvtonconcatenationneedvirtual,choi2024improving}. We further compare DEFT-VTON’s performances with existing SOTA models’ performances on complex multi-garment try-on tasks in Figure \ref{fig:SOTA}. DEFT-VTON achieves the best performance across different models and different tasks, further confirming its performance boosts. 

\subsection{Quantitative Results}
\subsubsection{Comparison with different model configurations} 
We first study which model configurations work best for the VTO task. 

We compare our approach with state-of-the-art methods. As shown in Table~\ref{tab:perform}, DEFT-VTON significantly outperforms the established VTO models across all metrics on the VITON-HD test dataset. DEFT-VTON achieves SOTA performances compared to existing works, while using less function evaluations. The improvement is significant, especially on the SSIM, FID, and the KID metric, showing that our model better maintains the structural information in the image. We show more comprehensive evaluations on VITON-HD test dataset in the Supplementary Material.
\begin{table}[ht]
\small
\begin{tabular}{lcccc}
\toprule
\multicolumn{1}{c}{\multirow{2}{*}{Models}} & \multicolumn{4}{c}{VITON-HD} \\ \cmidrule(lr){2-5} 
\multicolumn{1}{c}{} & \multicolumn{1}{l}{SSIM $\uparrow$} & \multicolumn{1}{l}{LPIPS $\downarrow$} & \multicolumn{1}{l}{FID $\downarrow$} & \multicolumn{1}{l}{KID $\downarrow$} \\ \midrule
StableVTON & 0.8543 & 0.0905 & 11.054 & 3.914 \\
LaDI-VTON & 0.8603 & 0.0733 & 14.648 & 8.754  \\
IDM-VTON & 0.8499 & 0.0603 & 9.842 & 1.123  \\
OOTDiffusion & 0.8187 & 0.0876 & 12.408 & 4.680 \\
CatVTON & 0.8704 & 0.0565 & 9.015 & 1.091  \\ \midrule
DEFT, $\mathcal{L}_{DSM}$, 25 steps & \underline{0.9118} & \underline{0.0533} & \textbf{8.3351} & \textbf{0.5212} \\
DEFT, $\mathcal{L}_{DSM}$, 15 steps & 0.9098 & \textbf{0.0521} & 8.6339 & 0.7916 \\
 \midrule
DEFT, $\mathcal{L}_{DSM}+\mathcal{L}_{CM}$, 25 st. & 0.9105 & 0.0564 & \underline{8.6310} & \underline{0.7243} \\
DEFT, $\mathcal{L}_{DSM}+\mathcal{L}_{CM}$, 15 st. & \textbf{0.9130} & 0.0542 & 8.8567 & 1.016 \\
\bottomrule
\end{tabular}
\caption{Results of baseline comparisons with DEFT-VTON \citep{morelli2023ladivtonlatentdiffusiontextualinversion,kim2024stableviton,choi2024improving,xu2024ootdiffusion,chong2024catvtonconcatenationneedvirtual}. Bold texts indicate best models, underlined texts indicate second best models. We provide results on the dresscode dataset in the appendix.}\label{tab:perform}
\end{table}

\subsubsection{Number of function evaluations}
We perform an ablation study of DEFT-VTON across different sampling steps on the VITONHD dataset. The evaluated scores stabalizes after 12 steps, highlighting consistency fine tuned DEFT-VTON's efficiency in the sampling process. We attach details to the ablation study in the Supplementary Material.

\subsection{Ablation study on adaptive coefficients}
\begin{figure}[hbt!]
    \centering
    \includegraphics[width=0.5\textwidth]{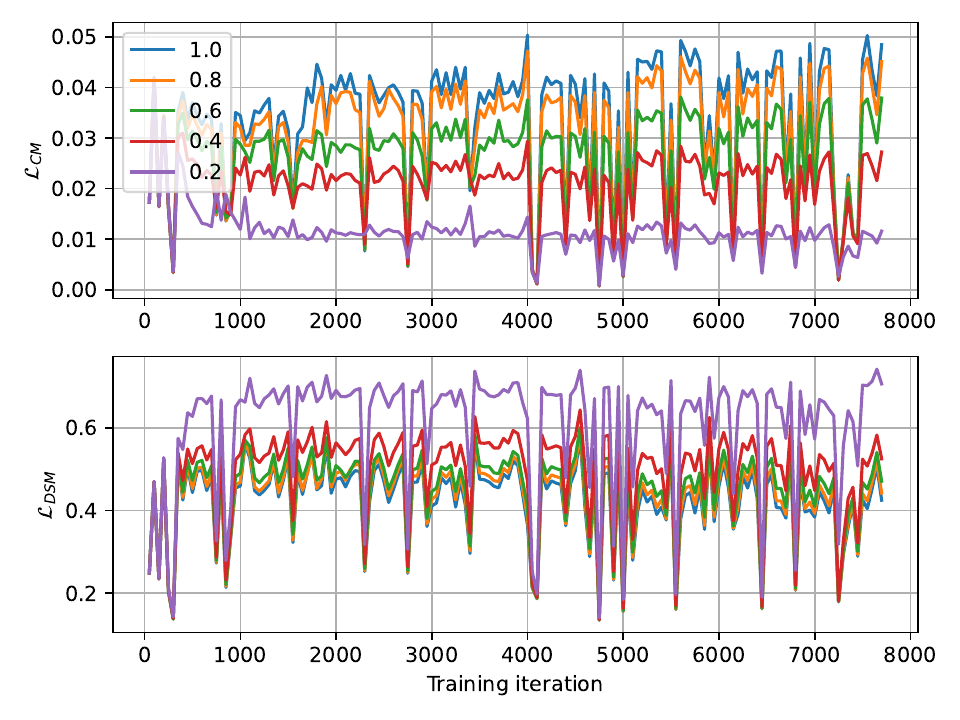}
    \caption{$\mathcal{L}_{CM}$ and $\mathcal{L}_{DSM}$ corresponding to different adaptive training coefficients. We use the same threshold for triggering a higher focus on $\mathcal{L}_{DSM}$ and ablate its coefficients.}
    \label{fig:cmDsm}
\end{figure}

We perform an ablation study on the adaptive balancing coefficients between the h-transform loss and the consistency loss. We use the same threshold for triggering a higher focus on $\mathcal{L}_{DSM}$ and the same coefficients for $\mathcal{L}_{CM}$, and ablate to study the impact of the coefficients of the coefficients for $\mathcal{L}_{DSM}$. As shown in Figure \ref{fig:cmDsm}, consistency loss increases and DSM loss decreases as we increase the coefficient. We find setting the coefficient at $0.6$ to achieve the best balance between $\mathcal{L}_{CM}$ and $\mathcal{L}_{DSM}$.

 \section{Conclusion}
\label{sec:conclusion}

In this paper, using a pre-trained diffusion transformer as backbone, we explore Doob's h-transform efficient finetuning (DEFT) and consistency training for virtual try-on tasks. Compared to existing VTO frameworks, the proposed DEFT-VTON completely freezes the backbone network, and trains an auxilary network with number of parameters as low as ~$1.42$\% of the backbone network. Being able to train auxiliary networks allows for efficient adaptations to ever-improving SOTA unconditional models regardless of their model sizes. To further accelerate the sampling process while preserving the VTO ability, we explore an adaptive balancing between the DEFT loss and the consistency loss. Ablation study shows DEFT finetuning leads to SOTA VTO performances, while the consistency finetuning accelerates the sampling process for up to $40$\% while retaining the same performances. Our model outperforms existing SOTA models both qualitatively and quantitatively, while requiring much less training cost. 

 \section{Limitation}
\label{sec:limitation}
While our proposed DEFT-VTON framework exhibits SOTA performances, as the h-transform network is only ~1.42\% of the pre-trained network, it relies heavily on the pre-trained unconditional model. The model might suffer from the same limitations as the base model, with any artifact or failure pattern observed in the unconditional model generally propagating into the DEFT-VTON model. 
 {
     \small
     \bibliographystyle{unsrt}
     \bibliography{main}
 }

 \clearpage
\setcounter{page}{1}
\maketitlesupplementary

\section{Ablation study}
\label{sec:ablation}
We provide ablation studies on the coefficients of $\mathcal{L}_{DSM}$ as well as the number of steps.

\begin{figure*}[t]
    \centering
    \includegraphics[width=0.93\textwidth]{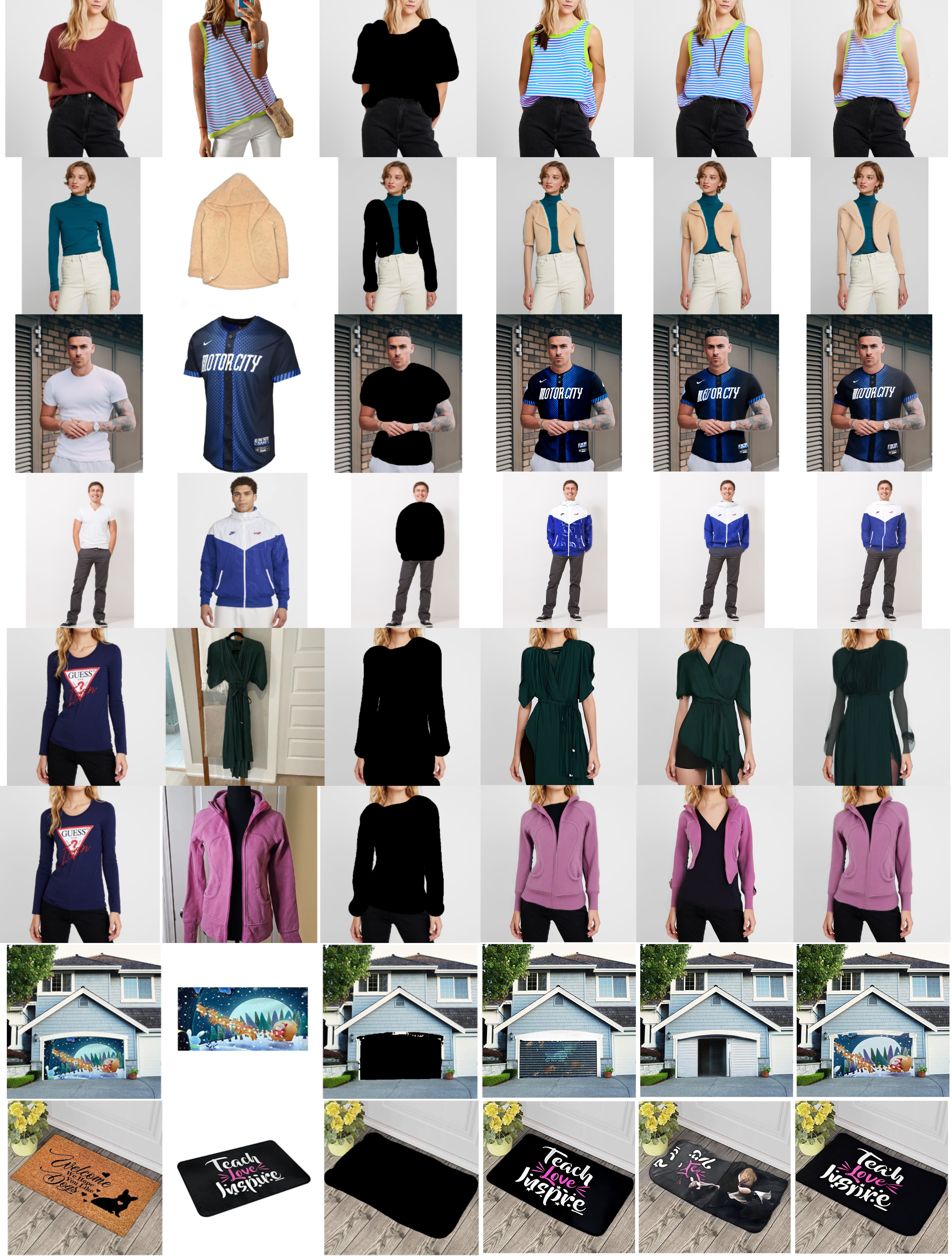}
    \caption{Results from test cases involving complex text and graphics preservation, unclear reference images, and unusual tasks. Each row shows a single task, starting with the original image and progressing through garment image, masked original image, 25-step sampling results (before consistency fine tuning), 15-step sampling results (before consistency fine tuning), and finally, 15-step sampling results (after consistency fine tuning).}
    \label{fig:ablation}
\end{figure*}

\paragraph{$\mathcal{L}_{DSM}$ coefficient}
While the table shows better SSIM and LPIPS scores for fewer steps (15) with changing $\mathcal{L}_{DSM}$ coefficients, the FID and KID scores, which assess perceptual quality, are worse. Empirically, we observe that the consistency finetuned models better preserve garment colors and complex text/graphics, as shown in Figure \ref{fig:ablation}, on some challenging tasks involving complex text/graphics preservation, unclear reference images, and tasks that are rare in the training dataset, 15 steps sampling of the consistency finetuned model qualitatively outperforms the one only finetuned with DEFT loss, reaffirming the quantitative observation that consistency finetuning improves sampling with fewer steps. 

We also observe that, when the coefficient on $\mathcal{L}_{DSM}$ is overly small, the DEFT-VTON model loses its VTO abilities.

\paragraph{Number of sampling steps}

Table \ref{tab:ablation} shows that the SSIM and LPIPS scores exhibit an initial rise and subsequent decline as sampling steps increase in the consistency fine-tuned DEFT-VTON model, implying an optimal performance at approximately 15 steps. We also observe that the FID and KID score consistently improve as we increase the number of sampling steps. Although this shows improvements in human perception, it does not always translate to better VTO results, with rising cases of hallucinations.


\begin{table*}[ht!]
\centering
\small
\begin{tabular}{lcccc}
\toprule
\multicolumn{1}{c}{\multirow{2}{*}{Models}} & \multicolumn{4}{c}{VITON-HD} \\ \cmidrule(lr){2-5} 
\multicolumn{1}{c}{} & \multicolumn{1}{l}{SSIM $\uparrow$} & \multicolumn{1}{l}{LPIPS $\downarrow$} & \multicolumn{1}{l}{FID $\downarrow$} & \multicolumn{1}{l}{KID $\downarrow$} \\ \midrule
StableVTON & 0.8543 & 0.0905 & 11.054 & 3.914 \\
LaDI-VTON & 0.8603 & 0.0733 & 14.648 & 8.754  \\
IDM-VTON & 0.8499 & 0.0603 & 9.842 & 1.123  \\
OOTDiffusion & 0.8187 & 0.0876 & 12.408 & 4.680 \\
CatVTON & 0.8704 & 0.0565 & 9.015 & 1.091  \\ \midrule
DEFT, $\mathcal{L}_{DSM}$, 25 steps & 0.9118 & \underline{0.0533} & \textbf{8.3351} & \textbf{0.5212} \\
DEFT, $\mathcal{L}_{DSM}$, 15 steps & 0.9098 & \textbf{0.0521} & 8.6339 & 0.7916 \\
 \midrule
DEFT, $0.2\mathcal{L}_{DSM}+\mathcal{L}_{CM}$, 12 steps & 0.9063 & 0.0782 & 25.7948 & 15.52 \\
DEFT, $0.2\mathcal{L}_{DSM}+\mathcal{L}_{CM}$, 15 steps & 0.9064 & 0.0798 & 27.8843 & 18.92 \\
DEFT, $0.2\mathcal{L}_{DSM}+\mathcal{L}_{CM}$, 25 steps & 0.9034 & 0.0853 & 33.2223 & 24.53 \\
\midrule
DEFT, $0.4\mathcal{L}_{DSM}+\mathcal{L}_{CM}$, 12 steps & \underline{0.9135} & 0.0553 & 9.1671 & 1.2612 \\
DEFT, $0.4\mathcal{L}_{DSM}+\mathcal{L}_{CM}$, 15 steps & \textbf{0.9136} & 0.0552 & 8.7922 & 0.9970 \\
DEFT, $0.4\mathcal{L}_{DSM}+\mathcal{L}_{CM}$, 25 steps & 0.9098 & 0.0581 & \underline{8.4704} & 0.6649 \\
\midrule
DEFT, $0.6\mathcal{L}_{DSM}+\mathcal{L}_{CM}$, 12 steps & 0.9122 & 0.0560 & 9.0136 & 1.1998 \\
DEFT, $0.6\mathcal{L}_{DSM}+\mathcal{L}_{CM}$, 15 steps & 0.9132 & 0.0553 & 8.6611 & 0.9104 \\
DEFT, $0.6\mathcal{L}_{DSM}+\mathcal{L}_{CM}$, 25 steps & 0.9100 & 0.0579 & 8.5988 & 0.6713 \\
\midrule
DEFT, $0.8\mathcal{L}_{DSM}+\mathcal{L}_{CM}$, 12 steps & 0.9112 & 0.0571 & 9.0765 & 1.2336 \\
DEFT, $0.8\mathcal{L}_{DSM}+\mathcal{L}_{CM}$, 15 steps & 0.9126 & 0.0561 & 8.7394 & 0.9378 \\
DEFT, $0.8\mathcal{L}_{DSM}+\mathcal{L}_{CM}$, 25 steps & 0.9101 & 0.0592 & 8.4160 & \underline{0.5474} \\
\midrule
DEFT, $1.0\mathcal{L}_{DSM}+\mathcal{L}_{CM}$, 10 steps & 0.8935 & 0.0862 & 12.7324 & 3.5322 \\
DEFT, $1.0\mathcal{L}_{DSM}+\mathcal{L}_{CM}$, 11 steps & 0.9111 & 0.0573 & 9.2023 & 1.3684 \\
DEFT, $1.0\mathcal{L}_{DSM}+\mathcal{L}_{CM}$, 12 steps & 0.9114 & 0.0571 & 8.9859 & 1.1134 \\
DEFT, $1.0\mathcal{L}_{DSM}+\mathcal{L}_{CM}$, 15 steps & 0.9125 & 0.0561 & 8.7113 & 0.8937 \\
DEFT, $1.0\mathcal{L}_{DSM}+\mathcal{L}_{CM}$, 20 steps & 0.9081 & 0.0617 & 8.8301 & 0.8030 \\
DEFT, $1.0\mathcal{L}_{DSM}+\mathcal{L}_{CM}$, 25 steps & 0.9091 & 0.0599 & 8.5058 & 0.5952 \\
\bottomrule
\end{tabular}
\caption{Results of baseline comparisons with DEFT-VTON on VITON-HD dataset. Bold texts indicate best models, underlined texts indicate second best models. }
\label{tab:ablation}
\end{table*}

\end{document}